\documentclass[preprint]{elsarticle}

\usepackage{balance} 

\begin{document}

\title{A Thermal Machine Learning Solver For Chip Simulation}

\author[inst1]{Rishikesh Ranade}

\affiliation[inst1]{organization={Ansys Inc, CTO Office},%Department and Organization
            addressline={2400 Ansys Dr.}, 
            city={Canonsburg},
            postcode={15317}, 
            state={PA},
            country={USA}}

\affiliation[inst2]{organization={Ansys Inc, CTO Office},%Department and Organization
            addressline={2645 Zanker Rd.}, 
            city={San Jose},
            postcode={95134}, 
            state={CA},
            country={USA}}
            
\affiliation[inst3]{organization={Ansys Inc, ESOBU},%Department and Organization
            addressline={2645 Zanker Rd.}, 
            city={San Jose},
            postcode={95134}, 
            state={CA},
            country={USA}}

\author[inst1]{Haiyang He}
\author[inst2]{Jay Pathak}

\author[inst3]{Norman Chang}
\author[inst3]{Akhilesh Kumar}
\author[inst3]{Jimin Wen}
% \email{jimin.wen@ansys.com}

% \affiliation{%

% \affiliation{%

% \renewcommand{\shortauthors}{Rishikesh Ranade et al.}

\begin{abstract}
  Thermal analysis provides deeper insights into electronic chips’ behavior under different temperature scenarios and enables faster design exploration. However, obtaining detailed and accurate thermal profile on chip is very time-consuming using FEM or CFD. Therefore, there is an urgent need for speeding up the on-chip thermal solution to address various system scenarios. In this paper, we propose a thermal machine-learning (ML) solver to speed-up thermal simulations of chips. The thermal ML-Solver is an extension of the recent novel approach, CoAEMLSim (Composable Autoencoder Machine Learning Simulator) with modifications to the solution algorithm to handle constant and distributed HTC. The proposed method is validated against commercial solvers, such as Ansys MAPDL, as well as a latest ML baseline, UNet, under different scenarios to demonstrate its enhanced accuracy, scalability, and generalizability. 
\end{abstract}


% \ccsdesc[500]{Applied Computing~Physical Sciences \& Engineering}

% \keywords{Machine learning, Chip Thermal Simulation, Autoencoders}

\maketitle

\section{Introduction}
It is well known that thermal issues can severely degrade the performance and reliability of chips \cite{sun2015localized, chandra2006s, mutschler2018new, emilio2019hybrid, kumar2022ml}. Overlarge peak temperatures and stiff thermal gradients can fatally impact transistor performance, stress, aging, electro-migration (EM), voltage drops and timing \cite{zhong2008thermal, peach2019protecting}. Hence accurate prediction of the maximum temperature and thermal gradient on the chip becomes important for the performance and reliability of chip-packaging systems used in several applications such as 5G, automobiles and computational hardware for Artificial Intelligence. Conventional Finite Element Analysis (FEA) or Computational Fluid Dynamics (CFD) based thermal analysis is computationally expensive due to the enormous system parameter space in the form of stiff powermaps and wide range of Heat Transfer Coefficients (HTCs), die thicknesses and chip sizes. As a result, batches of simulations are required to be solved from scratch every time new system parameters of electronic chips are considered. 

Recently, a multitude of machine learning methods have been proposed to enhance and accelerate physics based numerical solvers in the context of electronic chip simulations. For example, a Deep Neural Networks (DNN) based fast static thermal solver has been proposed in \citet{wen2020dnn} to generate a high-resolution Delta T map. In a new branch of study, researchers have employed Physics-Informed Neural Networks (PINNs) \cite{raissi2018hidden} to solve chip problems \cite{cai2021physics}. \citet{hennigh2021nvidia} use PINNs to predict temperature profiles resulting due to variations in the FGPA heat sink geometry for a uniform power map. Similarly, \citet{ranade2021discretizationnet} and \citet{he2020unsupervised} use discretization-based techniques in combination with neural networks to predict temperatures on chips for powermaps sampled from a Gaussian distribution. \citet{stipsitz2022approximating} presented a proof-of-concept approach to provide approximate predictions of steady-state temperatures using convolutional neural networks. \citet{chhabria2021thermal} use a convolutional encoder-decoder network to learn a mapping between powermap and temperature. Most studies in relation to thermal simulation of chips make simplifying assumptions about the power map or other system parameters such as HTC. The high dimensional parameter space involved in on-chip thermal simulations usually makes it challenging for conventional deep learning models to predict accurate temperatures for unseen input parameters.

In this paper, we propose a thermal ML-Solver, which is based on a recent work \cite{ranade2021composable}, for accurately simulating temperatures on electronic chips subjected to a wide range of system parameters. Our approach extends the previous method to modeling the complicated use case of electronic chips as well as proposes a slight modification to the solution algorithm to account for global system parameters as well as distributed HTCs. The input distributions considered in this paper have noticeable discontinuities and wider ranges of magnitudes and spatial distributions, which are extremely challenging to model. This paper demonstrates how the proposed approach can accurately model the temperature prediction widely ranging distributions of powermap and HTCs. The ability of this approach to accurately model such high-dimensional power maps and HTCs makes it unique and different from other ML approaches.

\textbf{Significant contributions}:
\begin{itemize}
    \item We propose a thermal ML-simulator based on the work by \citet{ranade2021composable} for electronic chip simulations over a wide range of parameters with varying spatial distributions.
    \item We modify the existing solution algorithm to handle global system parameters. This work can be considered as one of the first implementations of ML to handle distributed HTCs.
    \item Our approach entails solving a constrained equilibrium during evaluation, similar to traditional PDE solvers. This results in better stability, robustness and accuracy as opposed to blackbox ML methods.
\end{itemize}

\subsection{Chip Thermal Analysis}
\begin{figure}[h]
  \centering
  \includegraphics[width=1\linewidth]{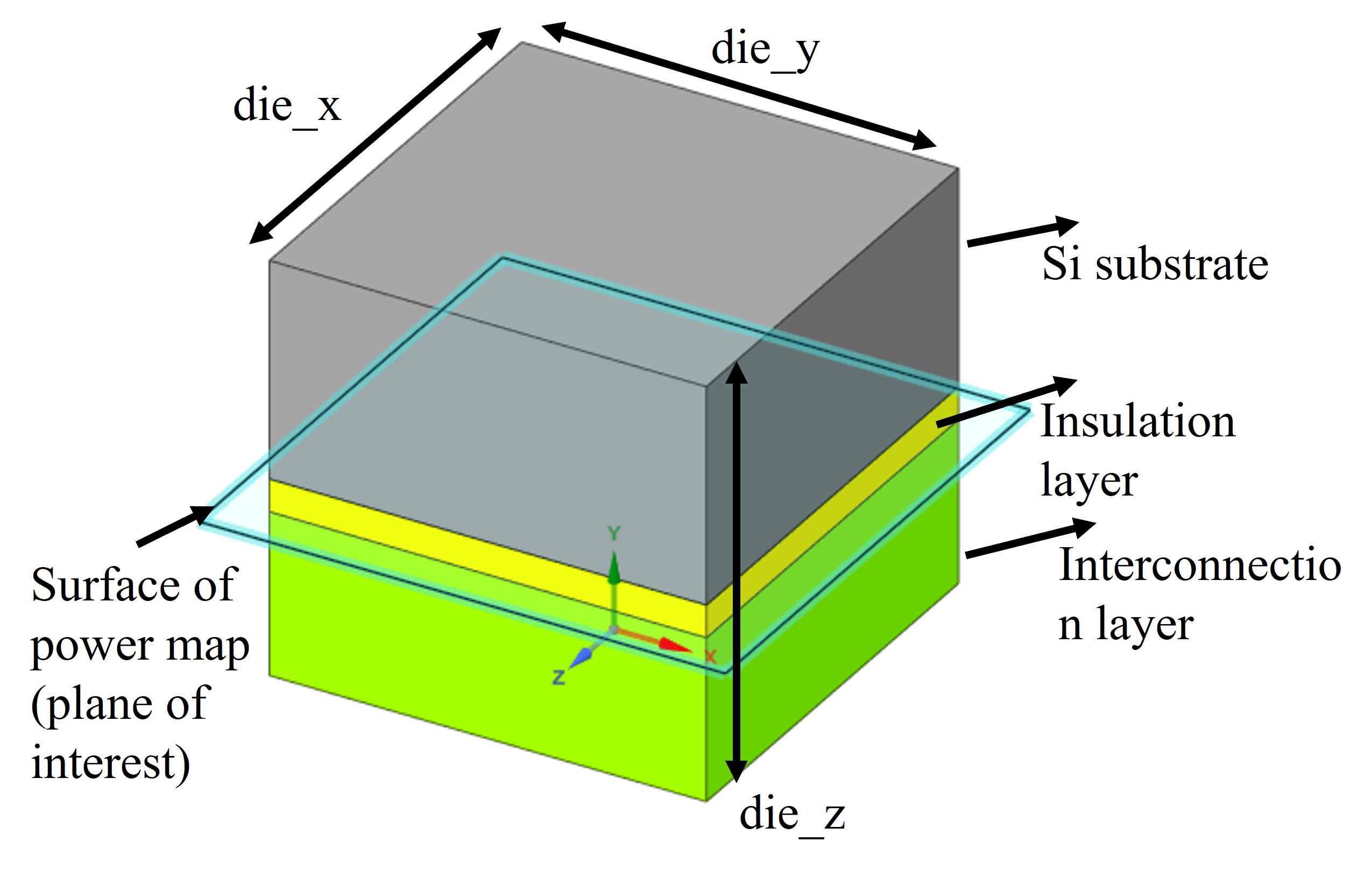}
  \caption{Schematic of the simplified three-layer chip model.}
  \label{chip_therm}
%   \Description{A woman and a girl in white dresses sit in an open car.}
\end{figure}

\begin{figure}[h]
  \centering
  \includegraphics[width=0.8\linewidth]{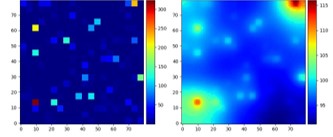}
  \caption{Example of powermap and temperature on chip.}
  \label{ex_power}
%   \Description{A woman and a girl in white dresses sit in an open car.}
\end{figure}

Chip thermal analysis may involve multiple components of the chip and various system parameters with high dimensionality and huge parameter space. This study focuses on a simplified three-layer chip model consisting of a Si substrate, an insulation layer and an interconnection layer. The schematic of the three-layer model is illustrated in Fig. \ref{chip_therm}. 
An arbitrary tile-based power map, which consists of rectangular regions of different heat source, each representing a different functional block on the chip, is sandwiched between the insulation layer and interconnection layer. For instance, a $4000 \mu m$ x $4000 \mu m$ chip with $200\mu m$x$200\mu m$ tile size will lead to a 20x20 array power map. An example of such power map is shown in Fig. \ref{ex_power}. Temperature distribution of the chip, especially the heating plane where the power map is applied, is of significant interest. In practice, such tile-based power could be random and incur drastic power gradient, which makes it challenging for conventional data-driven based approaches to learn, not to mention generalization to any unseen random power maps. Additionally, the temperature distribution on the heating plane is greatly affected by system parameters such as HTCs distribution on the boundary or die thickness, the effect of which is difficult to capture. Hence, in this study we only focus on accurately modeling chip temperature predictions on the heating plane across a wide range of parameters. In a system-level simulation, the three-layer die is cooled either by natural or forced convection. The cooling effect of air is modeled here using HTC distributions on top and bottom surfaces. 

\subsection{Data Generation} \label{data_gen}

In this paper, we present two use cases with varying levels of complexity. In the first case, the input parameters consist of a powermap distribution on the heating plane, HTCs on boundaries and die thickness. Constant HTC values are applied on the top and bottom boundaries of the die, whereas other boundaries have adiabatic boundary conditions. The HTC varies in between $1e^{-6}$ and $1e^{-5}$, while the die thickness varies between $20 \mu m$ and $200 \mu m$. $5000$ numerical simulations are carried out utilizing a high-fidelity numerical solver, Ansys MAPDL, for simulating the heat transfer on a $4000 \mu m$ chip with a grid resolution of $256$x$256$x$32$ for random power maps, heat transfer coefficients and Die thicknesses. Examples of the power map and the corresponding temperature on the chip are shown in Fig. \ref{ex_power}. 

The second use case presents a more challenging scenario close to industrial applications, where the HTCs specified on the top and bottom boundaries also have a spatial tile-based distribution similar to that of the power map and the tile size is smaller. The HTC distribution on each tile are varied randomly from $-1e^{-6}$ to $1e^{-6}$. Additionally, we reduce the tile size of powermap and HTC from 200 $\mu m$ down to $50 \mu m$. In this case, $2500$ numerical simulations are carried out using Ansys MAPDL for random power maps and HTCs.

In both cases, other parameters such as interconnection layer thickness, insulation layer thickness, Si substrate thermal conductivity, insulation layer thermal conductivity and interconnection layer conductivity are kept constant. The power on each tile is randomly sampled between $0$ and $300 mW$ and the total power on the chip is conserved among all the training samples. Even though the data is generated on a 3-D die, the solutions are retained only on the chip where the power map is applied. Hence, the temperature is predicted only on a surface but its spatial distribution and magnitude are affected by system parameters such as HTC and die thickness. As a result, these parameters are accounted for in the thermal ML-Solver along with power map. The division between the training and testing datasets is $80/20$.   
% \begin{table}
%   \caption{System parameters for 3-layer chip model}
%   \label{tab:data}
%   \begin{tabular}{cc}
%     \toprule
%     System Parameters&Value/Ranges\\
%     \midrule
%     Heat Transfer Coefficient $(mW/\mu m^2K$) & $1e^{-6}-5e^{-5}$\\
%     Die z $\mu m$ & $20$-$200$ \\
%   \bottomrule
% \end{tabular}
% \end{table}
\section{Thermal ML-Solver Details}

\begin{figure*}
  \centering
  \includegraphics[width=1\linewidth]{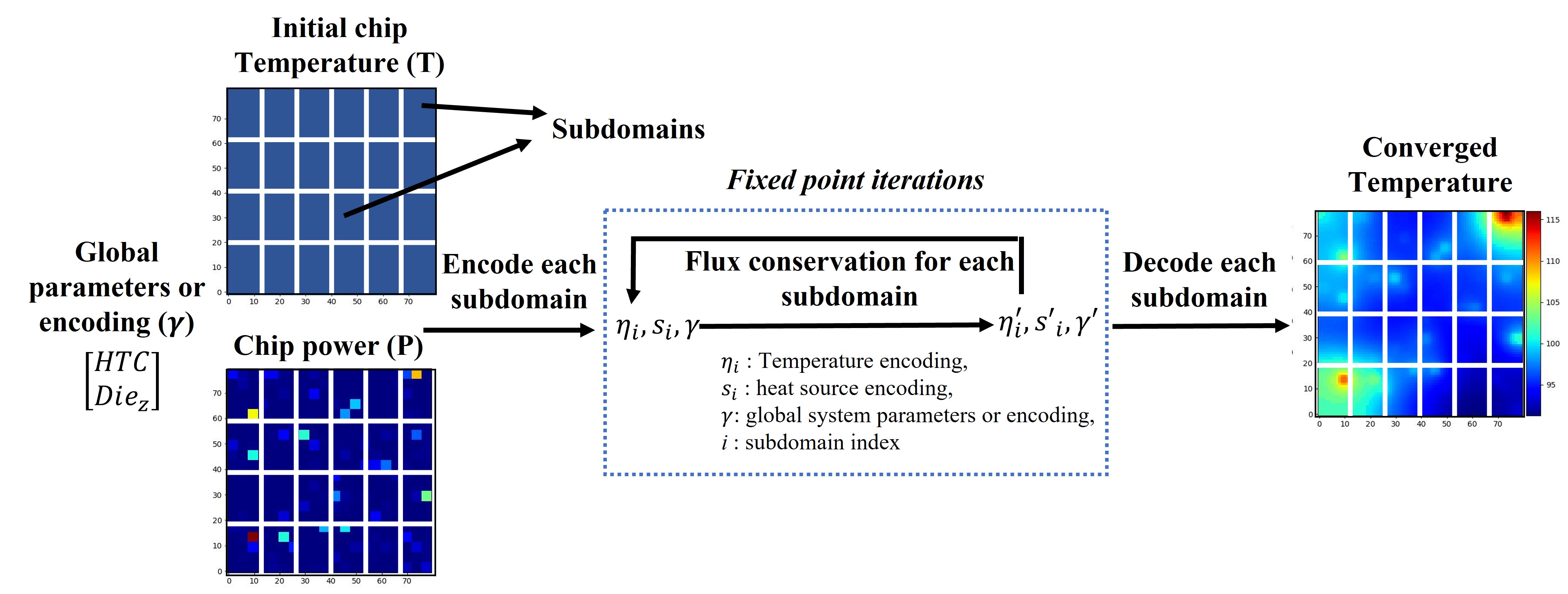}
  \caption{Solution algorithm}
  \label{comlsim-thermal}
\end{figure*}

A schematic of our model is described in Fig. \ref{comlsim-thermal}. Here we describe the components of our thermal ML-solver and the modifications to the existing algorithm found in the work by \citet{ranade2021composable}.
\begin{enumerate}
    \item \textbf{Decomposition of computational domain:} The computational domain corresponding to the 2-D chip (or plane of interest in Fig. \ref{chip_therm}) is decomposed into subdomains of equal physical sizes, where each subdomain consists of $m^2$ computational elements. As prescribed in \citet{ranade2021composable}, based on an extensive ablation study , the value of $m$ is set to $16$ resulting in around $256$ subdomains for the $4000 \mu m$ chip. 
    \item \textbf{Encoding on subdomains:} An initial temperature of $T=300 K$ and a given power map, $P$ are encoded into lower-dimensional encodings $\eta$ and $s$ on every subdomain using pre-trained encoders. The global system parameters, $\gamma = HTC, Die_z$, are kept constant across all the subdomains in the first use case. The conditioning of all the subdomains with global parameters improves the predictive capability of the solver. In the second use case, HTCs on top and bottom surfaces are decomposed in to subdomains and encoded similar to the powermap but $Die_z$ is used as a global parameter. Even though the HTC is applied on the top and bottom surface of the die, they are considered to be applied on the plane of the chip for the sake of modeling a 2-D problem.
    \item \textbf{Constrained Thermal solver:} A constrained equilibrium problem is solved using fixed point iterations where the encodings ($\eta, s, \gamma$) on each subdomain and its neighbors are evaluated iteratively using a pretrained flux conservation autoencoder. In each iteration, the output encoding corresponding to temperature, $\eta'$, evaluated by the flux conservation autoencoder, is used to replace the input temperature encoding, $\eta$ for the next iteration. Other encodings such as $s$ and $\gamma$ are kept fixed. The fixed encodings serve as hard constraints and steer the temperature encoding towards an equilibrium, which resembles the encoding of converged temperature. The fixed point iteration is stopped when the following condition is met, $L_2 (\eta - \eta') < 1e^{-8}$. 
    % This component of our algorithm closely resembles traditional PDE solvers. 
    % The inclusion of global parameters on each subdomain adds stability to the convergence of the solver.
    \item \textbf{Decoding on subdomains:} The converged temperature encodings on subdomains at the end of the fixed point iterations are decoded into temperatures and post-processed on the entire computational domain.
\end{enumerate}

It may be observed that our approach strongly draws ideas and inspiration from traditional PDE solvers. The novel iterative inferencing approach is designed to provide stability and robustness to the solution methodology in comparison to blackbox ML methods. Being the workhorse of our approach, the flux conservation autoencoder is useful in exchanging information amongst neighboring subdomains and from the boundary to the interior. This is similar to traditional solvers where flux conservation is employed between computational elements using numerical approximations for the same purpose. Finally, it is important to note that the training portion in this method corresponds to simply training $3$ autoencoders in use case 1 and $4$ in use case 2. Autoencoders are an unsupervised ML technique to obtain lower-dimensional encodings from higher-dimensional fields \cite{goodfellow2016deep}. It is important to note that we don't explicitly train a model to learn a mapping between the inputs (power map, HTC and thickness) and outputs (Temperature). Instead, we solve a constrained equilibrium problem in the lower-dimensional latent space during inference to predict the temperature for a fixed set of inputs, such as power map, HTC and die thickness, which are specified by the user. The implementation of this component is unique and differentiates our approach from common ML methods.

\subsection{Autoencoders in Thermal ML-Solver}
\begin{figure*}
  \centering
  \includegraphics[width=1\textwidth]{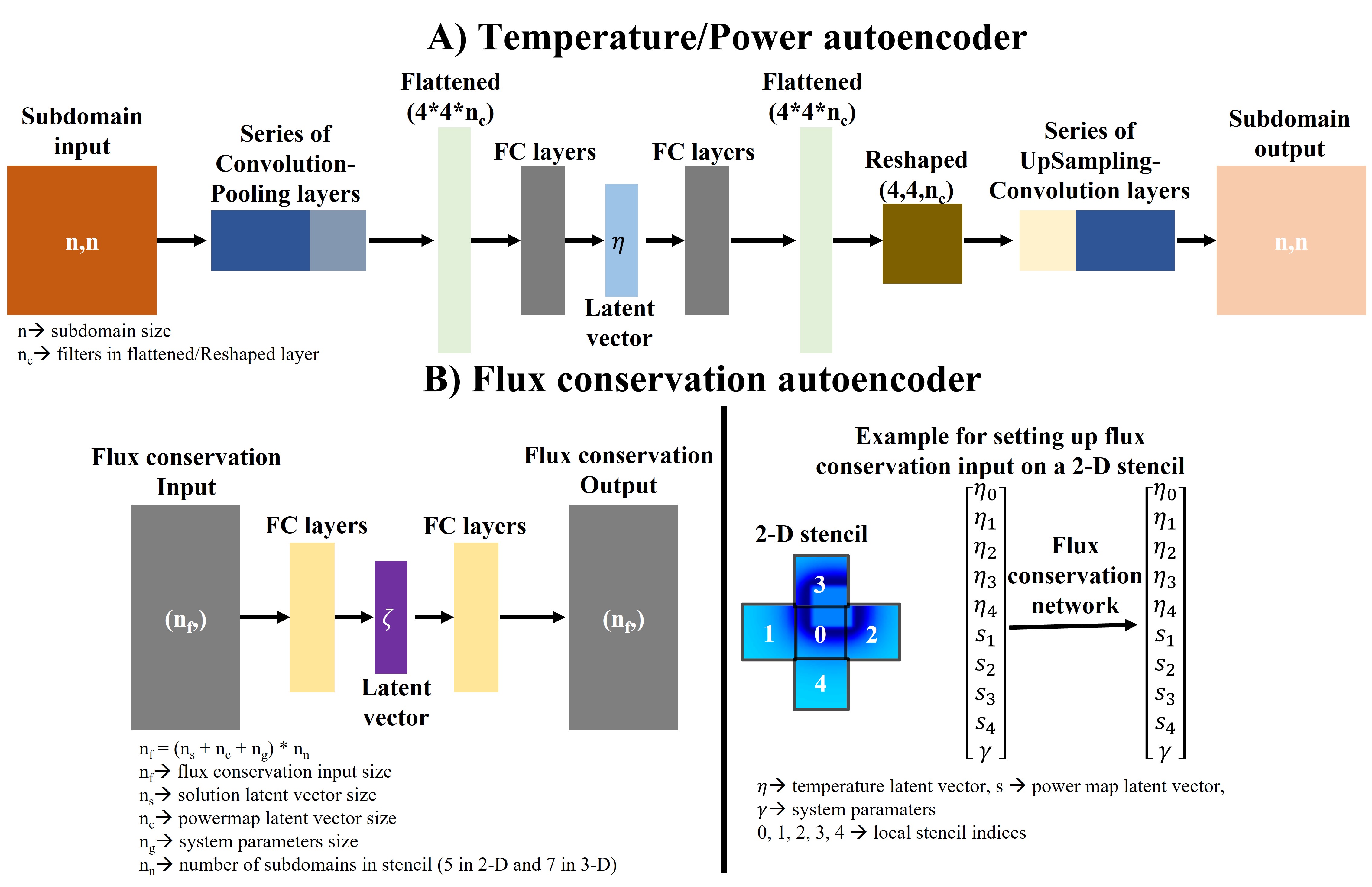}
  \caption{A) Temperature and power map autoencoder architecture and B) Flux conservation autoencoder architecture}
  \label{autoencoders}
\end{figure*}

In Figure \ref{autoencoders}, we describe the network architectures of autoencoders used in our algorithm. The autoencoders in Fig. \ref{autoencoders}A are CNN-based encoder-decoder networks. The encoder part of this network downsamples the subdomain field into a corresponding lower-dimensional encoding. The decoder part of the network upsamples the encoding back to the original input field. Separate autoencoders are trained for temperature and powermap for both use cases. Additionally, an autoencoder for distributed HTC is also trained in use case 2. On the other hand, the flux conservation autoencoders are DNN-based encoder-decoder networks. The temperature, powermap and system parameters on a 2-D stencil of $5$ subdomains are encoded using the pretrained encoders and stacked together as shown in Fig. \ref{autoencoders}B to form the input for this network. All the autoencoders are trained in Tensorflow 2.2.0 using an Adam optimizer with mean absolute error loss. We choose autoencoders over other compression techniques because non-linear autoencoders have powerful generalization and denoising capabilities with reasonable compression ratios \cite{goodfellow2016deep}. This improves the stability, robustness and ultimately the convergence of the constrained equilibrium solve employed during the evaluation of our approach.

\subsubsection{\textbf{Specific architecture details:}} 

\textbf{Solution and Condition autoencoder:} These are CNN based encoder-decoder networks. The encoder part of the network has a series of 3 convolution layers followed by max pooling. The number of filters in the convolution layers are 16, 32 and 64 respectively. The output of the convolution layer is flattened and passed through 2 dense layers of size 1024 and 21, where 21 is the size of the latent vector. In this work, the solution, power map and distributed HTC fields are encoded to the same latent size. The decoder part of the network is exactly symmetric to the encoder.

\textbf{Flux conservation autoencoder:} The flux conservation autoencoder has an input size of $230$ in the case of constant HTC and die thickness and $330$ with distributed HTC. This is a fully-connected encoder-decoder based network. The encoder part of the network has 2 hidden layers with sizes 1024 and 512. The latent vector for this network has a size of 35. The decoder part of the network is exactly symmetric to the encoder.

\section{Results and Discussion}

In this section, we present results obtained from our approach for unseen test cases across both the use cases and compare it with Ansys MAPDL. For the first use case with constant HTC, we provide additional experiments to demonstrate the generalizability of our approach for varying chip sizes as well as tile size and out-of-range HTC values. Finally, we compare our solver with a popular ML network, UNet \cite{ronneberger2015u}, across all the experiments. It must be noted that different instances of the solver are trained for the 2 use cases due to the differences in HTC inputs.

\subsection{Use case 1: Constant HTC}

\subsubsection{\textbf{Unseen test cases}} \label{unseen}
\begin{figure}[h!]
  \centering
  \includegraphics[width=1\linewidth]{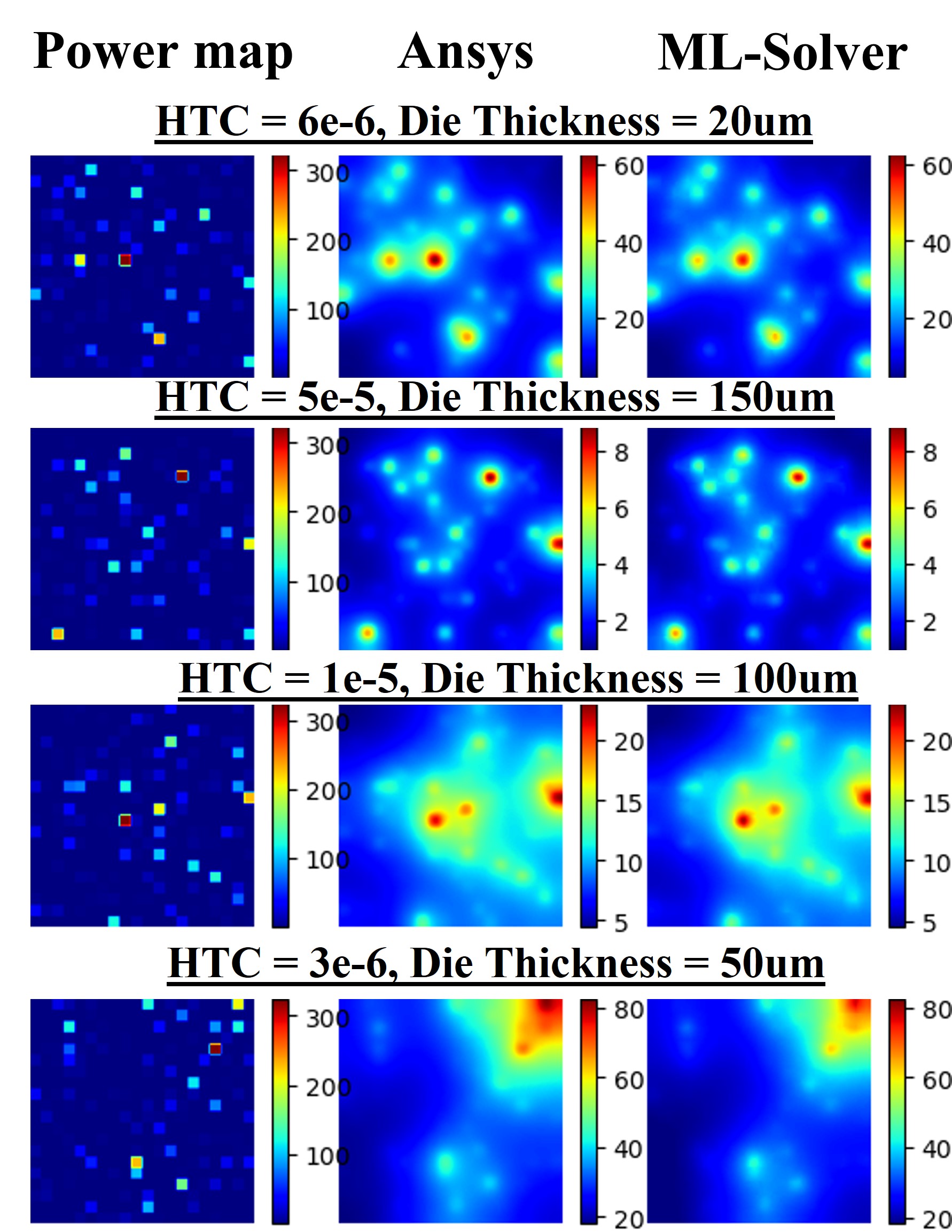}
  \caption{Use case 1: Comparison of ML-Solver with Ansys Mechanical.}
  \label{test1}
\end{figure}

In this experiment we test the thermal ML-Solver for unseen, randomly sampled powermaps, HTC and die thickness. The testing is carried out on $500$ samples not included in the training set. The relative $L_2$ error, defined as $\epsilon = \frac{L_2 (T_{true} - T_{pred})}{L_2(T_{true})}$, is $0.078$. The relative $L_2$ error serves as the most suitable metric in this case because the system parameters can result in a wide range of chip temperature magnitudes. 
Randomly selected samples are plotted in Fig. \ref{test1} and show that the thermal ML-Solver matches well with results obtained from Ansys MAPDL. 

\begin{figure}[h!]
  \centering
  \includegraphics[width=1\linewidth]{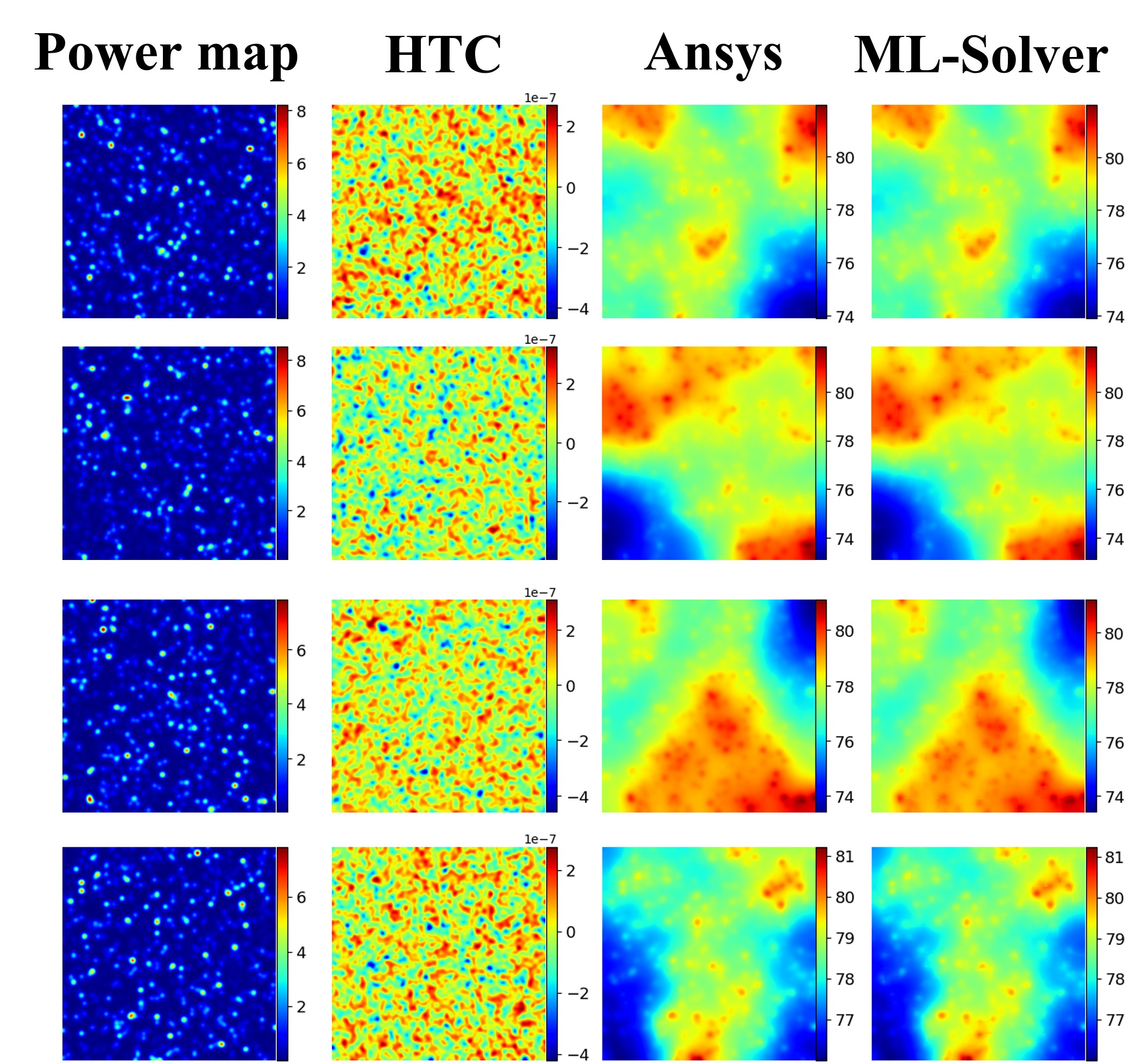}
  \caption{Use case 2: Comparison for distributed HTC on top and bottom die surfaces and 50 $\mu m$ tile size with Ansys Mechanical.}
  \label{test5}
\end{figure}

\subsubsection{\textbf{Generalization to larger chips}}\label{chipsize}

\begin{figure*}[h!]
  \centering
  \includegraphics[width=1\linewidth]{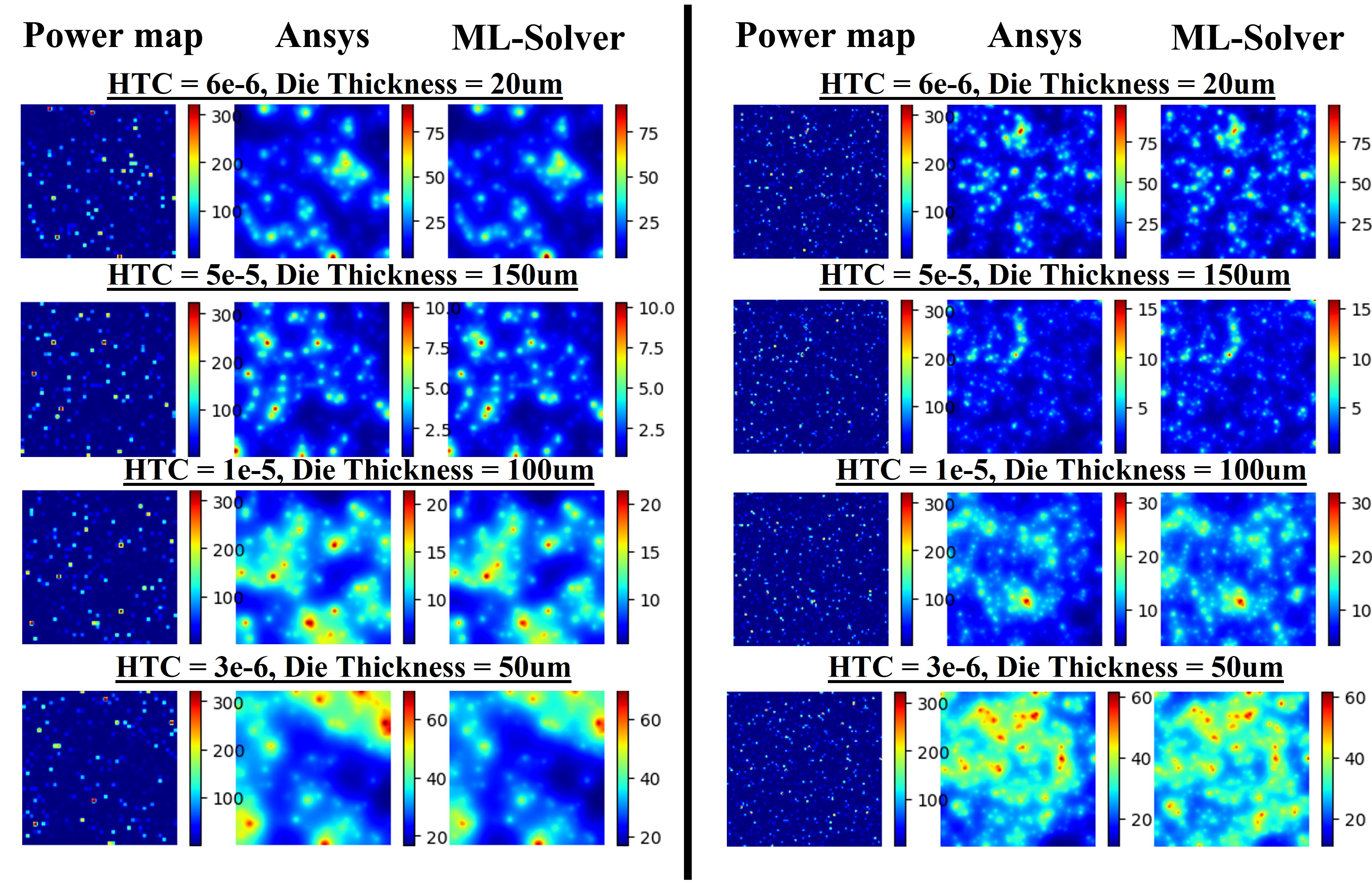}
  \caption{Use case 1: Comparison for bigger chip sizes 8000$\mu m$ (left) and 16000$\mu m$ (right) with Ansys Mechanical.}
  \label{test2}
%   \Description{A woman and a girl in white dresses sit in an open car.}
\end{figure*}

In this experiment, we test the generalizability of our approach to chips of size $8000 \mu m$ and $16000 \mu m$. It should be noted that all models are trained on a chip of physical size of $4000 \mu m$. The testing is carried out on $5$ testing samples for each chip size with randomly varying powermaps, HTC and die thickness. The relative L-2 error for the $8000 \mu m$ and $16000 \mu m$ chips are $0.058$ and $0.089$ respectively. The prediction accuracy of our thermal solver to larger chips may be attributed to the local learning approaches coupled with iterative inferencing schemes. Additionally, in Fig. \ref{test2} we compare the contour plots of our approach with Ansys MAPDL for randomly selected testing samples. It may be observed that for both chip sizes, the thermal ML-Solver accurately models the temperature. Although not shown here, we expect our solver to scale to even larger chip sizes because of the local learning approach employed in our approach. 

\subsubsection{\textbf{Generalization to a different tile size}}\label{tilesize}
Next, we demonstrate generalization to a different tile size of the powermap equal to $250 \mu m$. It should be noted that all models are trained on tile sizes of $200 \mu m$. The testing is carried out on $5$ testing samples for each chip size with randomly varying powermaps, HTC and die thickness. The relative L-2 error obtained is $0.14$. It may be observed from the contour plots in Fig. \ref{test4} that the results match reasonably well with respect to Ansys MAPDL.
% \begin{figure}[h!]
%   \centering
%   \includegraphics[width=\linewidth]{pictures/test_case4.jpg}
%   \caption{Comparison between Thermal ML-Solver and Ansys APDL for power tile size 250$\mu m$.}
%   \label{test3}
% \end{figure}
\subsubsection{\textbf{Generalization to out-of-range HTCs}}\label{htc_thick}
\begin{figure*}[h]
  \centering
  \includegraphics[width=1\linewidth]{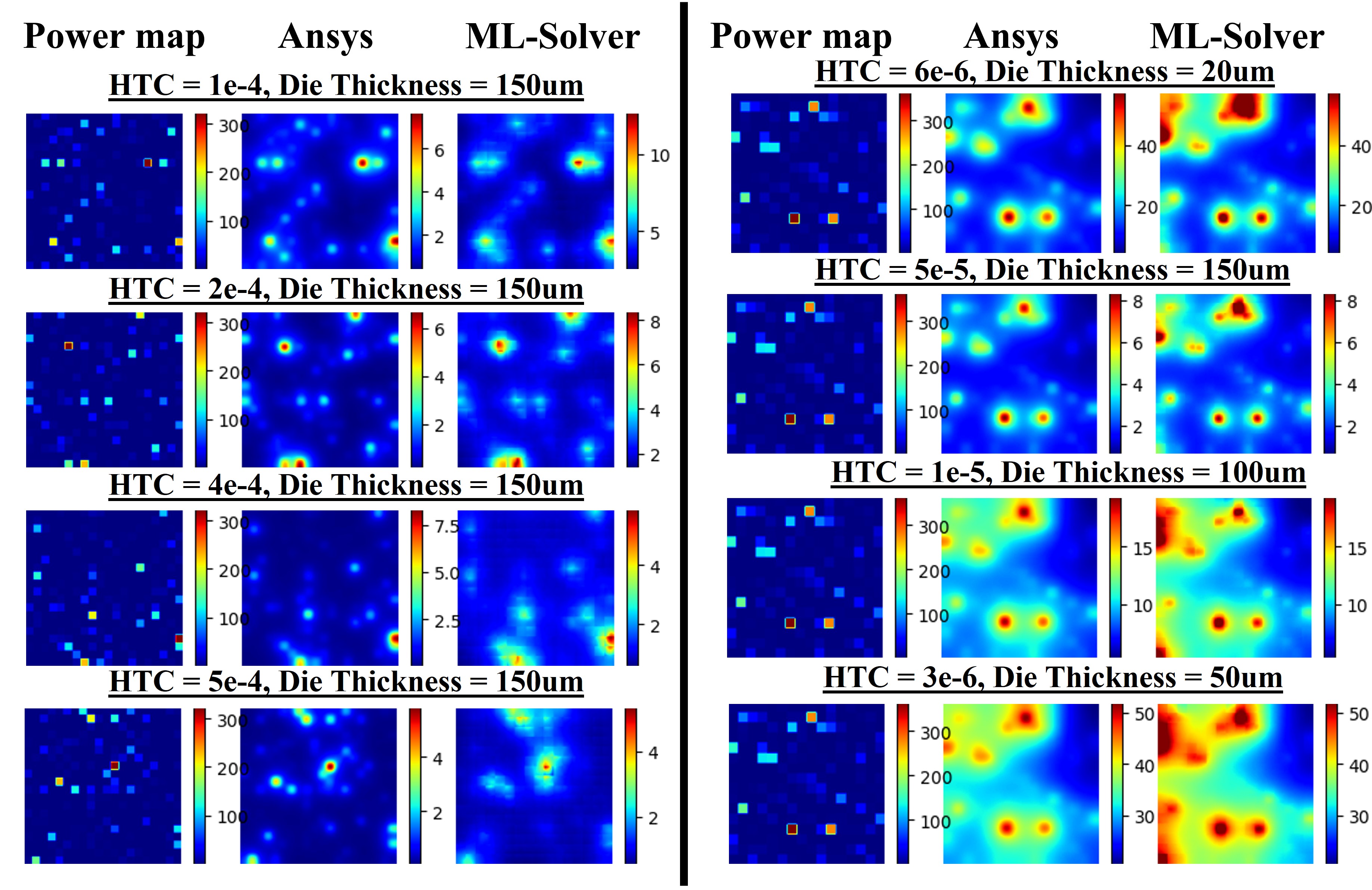}
  \caption{Use case 1: Comparison for higher HTC values (left) and bigger power map tile sizes (right) with Ansys Mechanical.}
  \label{test4}
\end{figure*}
In this section, we test the performance of the thermal ML-Solver in the extrapolation regime of HTC. We generate $5$ samples for HTC $1e^{-4}$, $2e^{-4}$, $3e^{-4}$, $4e^{-4}$ and $5e^{-4}$ for a die thickness of $150\mu m$ and random power distribution. The average relative $L_2$ error is $0.098$. It may be observed from the contour plots in Fig. \ref{test4} that the results match reasonably well with respect to Ansys MAPDL and the error progressively increases as we go further away from the training HTCs.

\subsection{Use case 2: Distributed HTC and 50 $\mu m$ tile} \label{dist_htc}

\subsubsection{\textbf{Unseen test cases:}}
In this section, we test the thermal solver on $250$ testing samples with unseen HTC and power map spatial distributions. Distributed HTCs can have a greater impact on temperature distribution making it more challenging to model. Additionally, the smaller tile sizes results in more local heating and stiffer peaks in temperature which are also difficult to model. Despite the challenges, our approach performs reasonably well with average relative L-2 errors of $0.045$ across all samples. It may also be observed from the contour plots in Fig. \ref{test5} that the temperature predictions between our thermal ML-Solver and Ansys MAPDL match well. Although not shown here, but our solver continues to generalize to bigger chip sizes for this use case as well with the same accuracy as the previous case.

\subsection{Comparison with other ML baselines}

Next, in table \ref{tab:baselines}, we report the relative $L_2$ comparisons between our approach and another popular approach, UNet \cite{ronneberger2015u} for all the test cases considered in this paper. The significantly better accuracy of results observed in table \ref{tab:baselines} demonstrates the superior generalization capability of our approach. Moreover, since our approach operates like a solver it can successfully scale to bigger chip sizes where traditional ML approaches cannot.

\begin{table}
  \caption{Comparison between thermal ML-Solver and UNet}
  \label{tab:baselines}
  \centering
  \begin{tabular}{||c|c|c||}
    % \toprule
    \hline
    Experiment&Thermal ML-Solver&UNet\\
    % \midrule
    \hline
    Section \ref{unseen} & $0.042$ & $0.049$\\
    \hline
    Section \ref{chipsize} ($8000\mu m$) & $0.058$ & $1.05$\\
    \hline
    Section \ref{chipsize} ($16000\mu m$) & $0.089$ & $1.18$\\
    \hline
    Section \ref{tilesize} ($250\mu m$) & $0.14$ & $0.2$\\
    \hline
    Section \ref{htc_thick} & $0.098$ & $0.35$\\
    \hline
    % Section \ref{dist_htc} & $0.045$ & $0.15$ \\
  % \bottomrule
\end{tabular}
\end{table}

\subsection{Computational time comparison}

The computational time required by MAPDL on a $4000 \mu m$ chip is about $30$ min on a single CPU. On the other hand, the thermal ML-Solver converges in less than $10$ seconds on a single CPU. Moreover, since our approach has similarities with traditional solvers it can be easily scaled to multiple CPUs and GPUs. The savings in time increases proportionally as the chip size increases. Finally, it must be noted that our approach employs iterative evaluation and can be slower than black-box ML models but it compensates by providing better accuracy, generalizability and scalability. For example, UNet requires less than 1 second to compute the solution but has a degraded accuracy and generalizability.

\section{Conclusion}
In this work, we introduce a thermal ML-Solver, which is based on the recently proposed CoAEMLSim approach \cite{ranade2021composable}, to accurately predict temperature for electronic chips across the high-dimensional system parameters in the form of power map, HTC, die thickness etc. In this paper, we show two use cases with constant and distributed HTCs, and demonstrate the accuracy of temperature predictions with respect to Ansys MAPDL on unseen test cases. Moreover, we provide additional experiments to test the generalizability of our approach across varying chip sizes, tile sizes and out-of-range HTCs and demonstrate our superior performance in comparison to a state-of the-art ML baseline. Although the thermal solver demonstrated in this work is trained and tested for powermap tile sizes of $200 \mu m$ x $200 \mu m$, the same approach can be easily extended to smaller tile sizes ranging from of $1-10 \mu m$. In future, we would like to extend the approach to transient chip problems and chip packages with geometric complexities and material variations.

% \begin{acks}
% To Prith Banerjee, Chief Technology Officer at Ansys Inc., for supporting this research.
% \end{acks}

\bibliographystyle{ACM-Reference-Format}
\balance
\bibliography{references}

% \appendix









\end{document}